\documentclass[runningheads]{llncs}
\usepackage{graphicx}
\usepackage{subcaption}
\usepackage{booktabs}
\usepackage{caption}
\usepackage{tabularx}
\usepackage{makecell}
\usepackage{array}
\usepackage{amsmath}
\usepackage{siunitx}
\usepackage{comment}

\begin{document}
\title{Nonverbal Immediacy Analysis in Education: A Multimodal Computational Model}
%
%
\author{Uroš Petković\inst{1, 3} \and
Jonas Frenkel\inst{2,3} \and
Olaf Hellwich\inst{1, 3} \and
Rebecca Lazarides\inst{2, 3}}
\authorrunning{U. Petković et al.}
%
\institute{Computer Vision \& Remote Sensing, Technische Universität Berlin, Germany \and
Department of Educational Sciences, University of Potsdam, Germany \and
Science of Intelligence (SCIoI), Cluster of Excellence, Berlin, Germany\\}
\titlerunning{ }
\maketitle              

\begin{abstract}
This paper introduces a novel computational approach for analyzing nonverbal social behavior in educational settings. Integrating multimodal behavioral cues, including facial expressions, gesture intensity, and spatial dynamics, the model assesses the nonverbal immediacy (NVI) of teachers from RGB classroom videos. A dataset of 400 30-second video segments from German classrooms was constructed for model training and validation. The gesture intensity regressor achieved a correlation of 0.84, the perceived distance regressor 0.55, and the NVI model 0.44 with median human ratings. The model demonstrates the potential to provide a valuable support in nonverbal behavior assessment, approximating the accuracy of individual human raters. Validated against both questionnaire data and trained observer ratings, our models show moderate to strong correlations with relevant educational outcomes, indicating their efficacy in reflecting effective teaching behaviors. This research advances the objective assessment of nonverbal communication behaviors, opening new pathways for educational research.

\end{abstract}
\section{Introduction}
The way in which instructors communicate is crucial for successful knowledge transfer. While traditionally, research has often focused on verbal communication as the primary mode of interaction in educational settings, empirical evidence highlights the critical role of instructors' nonverbal behaviors in promoting cognitive, affective, and motivational learning ~\cite{witt2004meta}.

The field of computer science, particularly within the domains of Social Signal Processing~\cite{vinciarelli2009social}, Affective Computing~\cite{rouast2019deep}, Educational Technology~\cite{mangal2019essentials}, and Human-Robot Interaction (HRI)~\cite{kennedy2017nonverbal}, has increasingly recognized the substantial influence of nonverbal communication in educational contexts. Advanced computational models and machine learning techniques offers novel avenues, that may hold distinct advantages over traditional observer or questionnaire assessments of nonverbal behaviors. Computer vision-based approaches, for instance, could provide increased objectivity and consistency by applying uniform criteria across diverse instances, mitigating subjective human biases ~\cite{vinciarelli2009social}. Additionally, these techniques allow for unobtrusive observations, preserving the authenticity of the educational setting by minimizing observer effects~\cite{gupta2018cvucams}. The scalability and efficiency of such systems facilitate the processing of large volumes of video data, while supporting cost-effective longitudinal studies, thereby aiding the evaluation of educational interventions~\cite{murphy2012machine}. 
Yet, despite the considerable advances, the challenge of capturing the complex and interactive dynamics of nonverbal signals remains significant. Existing research has predominantly adopted a reductive approach, focusing on specific nonverbal cues such as pointing or gaze, often analyzing them in isolation from each other~\cite{ahuja2019edusense,bosch2018quantifying}. The nature of nonverbal communication, however, is characterized by the intricate interaction of multiple cues that function collectively. The effects of individual cues are not merely additive; rather, the combination of different nonverbal cues can alter their individual intensities and inherent meanings~\cite{patterson2014reflections}. As social cues represent a unified percept, they need be assessed holistically~\cite{kennedy2017nonverbal,zaki2013cue}, necessitating an integrated approach to studying the collective effects of nonverbal behaviors alongside individual cues.

A construct that effectively describes nonverbal behavior in such a manner is that of nonverbal immediacy (NVI)~\cite{mehrabian1968some}. Closely related to the construct of enthusiasm, NVI comprises behaviors that transmit positive signals, such as sympathy and warmth, effectively reducing the psychological distance between instructors and students. Common behaviors associated with NVI include gestures, eye contact, a relaxed posture, and smiling~\cite{liu2021does}. Meta-analyses have consistently demonstrated positive correlations between NVI and various cognitive and affective-motivational learning outcomes ~\cite{witt2004meta,liu2021does}.

Against this background, this paper presents a novel computer vision-based approach for estimating NVI scores of teachers by integrating multiple dimensions of nonverbal cues. Our research aims to develop computational models that analyze facial expressions, gesture intensity, and spatial dynamics collectively. By leveraging machine learning techniques, we aim to provide a more comprehensive understanding of nonverbal communication in an educational context.

Our main contributions in this paper are as follows: a) We present the first known effort to estimate NVI computationally from video recordings. b) We develop a gesture intensity measurement model that captures a continuous spectrum of gestures, moving beyond binary recognition to more subtly distinguish variations in gesture intensity. c) Our research introduces the first model for assessing perceived distance as a key aspect of spatial dynamics in educational settings. d) We have developed a specialized dataset, labeled by trained raters, for the training and validation of models in NVI, perceived distance, and gesture intensity.

\section{Related Work}
\label{sec:relatedwork}
This section presents a review of studies on teachers' behaviors in educational settings, categorized into two main areas: 1) individual behavior recognition identifies specific actions and movements, such as hand raises and body poses, and 2) overall teaching behavior assessment evaluates teachers' performance by analyzing broader behavioral patterns and interactions, providing insights into teaching quality and classroom dynamics.

\subsubsection{Individual Behavior Analysis}
In~\cite{ahuja2019edusense}, depth cameras extract skeletons for classroom monitoring, enhancing teacher development by analyzing activities. The models use these skeletons to identify behaviors like hand raises, body poses, and speech patterns via machine learning. The dataset includes 30 participants (5 instructors, 25 students), with 1545 body instances and 60 speech/silence audio instances. Challenges include the need for high-resolution cameras for precise skeleton extraction and adapting to diverse classroom layouts.

Similarly, TeachLivE~\cite{barmaki2015providing} tracks teachers' interactions with virtual 3D students using skeleton extraction, offering posture feedback. In a study with 34 trainee teachers, initial skeletal posture feedback led to improved body language later, showcasing real-time behavioral assessment's efficacy. However, TeachLivE's need for clear views for accurate tracking and extraction limits its use in typical classrooms.

The study presented in~\cite{bosch2018quantifying}, a video-based motion estimation model evaluates teachers' non-verbal behaviors in classrooms, including gesturing and walking. Analyzing nine lectures from a Canadian university, it utilizes motion detection, camera pan, and zoom techniques on 5415 30-second video segments. While effective for sustained activities like walking, accuracy for brief gestures is constrained by the 30-second analysis window and the limited dataset of nine teachers. Shorter time frames and a diverse dataset are necessary to enhance detection of transient movements.

In the research documented in~\cite{wu2020recognition}, videos of distinguished educators are analyzed for behaviors like blackboard-writing, questioning, displaying, instructing, describing, and non-gesture behavior. Using 3-second clips, the study combines RGB video and skeleton data, exploring early and late fusion methods. It stresses the necessity of detailed dataset information to improve model generalizability, highlighting the limited effectiveness of late fusion techniques here.

These single-behavior approaches focus primarily on action recognition using skeletal data. While this approach significantly reduces visual information and improves robustness, it also lacks crucial contextual details needed to estimate behaviors such as gesture intensity and perceived distance. To address these gaps, our proposed models retain visual information, allowing for a more comprehensive analysis. Furthermore, unlike previous studies that rely on depth cameras, our approach does not require such specialized equipment, thereby increasing its practicality for diverse educational settings.

\subsubsection{Overall Teaching Behavior Assessment}

In the study~\cite{sumer2018teachers}, a novel framework is proposed for analyzing teacher perceptions in classroom settings, utilizing a combination of egocentric video recordings and mobile eye tracker data. This method employs face detection and tracking technologies to meticulously monitor the teacher's focus on individual students, offering insights into attention patterns in relation to student characteristics such as gender. While this approach represents a significant advancement in educational research, enabling detailed attention maps and a deeper understanding of teacher-student interactions, it is constrained by the small size of the dataset and the challenges posed by variable recording conditions, such as motion and lighting variations.

In~\cite{chen2020research}, a model evaluates university instructors' teaching enthusiasm through sound feature extraction, facial expression recognition, and pose estimation. A cascading feature fusion method enhances predictive accuracy, with estimations analyzed using a BP neural network. Challenges arise in processing missing values, like substituting data when faces or body parts are obscured, potentially impacting prediction accuracy. The dataset includes 1004 10-second videos of 46 university teachers, offering a broad foundation for analyzing teaching behaviors across diverse course settings.

The study in \cite{ramakrishnan2021toward} used the ACORN system to assess the 'positive climate' in classrooms by analyzing 15-minute videos, longer than typical datasets. This included the UVA Toddler dataset with 300 videos and the Measures of Effective Teaching (MET) dataset~\cite{kane2013have}, with 5574 CLASS-coded segments~\cite{pianta2008classroom}. The MET dataset and the related experiment focused exclusively on auditory data. The system utilized auditory features, such as low-level acoustic signals, key phrase detection, and speech recognition, which made up the larger proportion of the input, along with visual features like facial expressions. Unlike our focus on predicting the NVI of teachers, their estimation of positive and negative climate included both teachers and students.

While there is a lack of studies specifically focused on measuring NVI, the broader field of automated teaching behavior assessment, which includes nonverbal elements, has seen some exploration. This field shows a significant variation in observation time windows, ranging from brief 10-second clips in~\cite{chen2020research} to more extended 15-minute sessions in~\cite{ramakrishnan2021toward}. These differences in approaches illustrate the challenges related to the duration of video analysis, as well as the diversity of sensors utilized, ranging from standard RGB cameras to advanced eye-tracking devices.

\begin{figure}[htbp]
    \centering
    \begin{subfigure}[b]{0.35\textwidth}
        \centering
        \includegraphics[width=\textwidth]{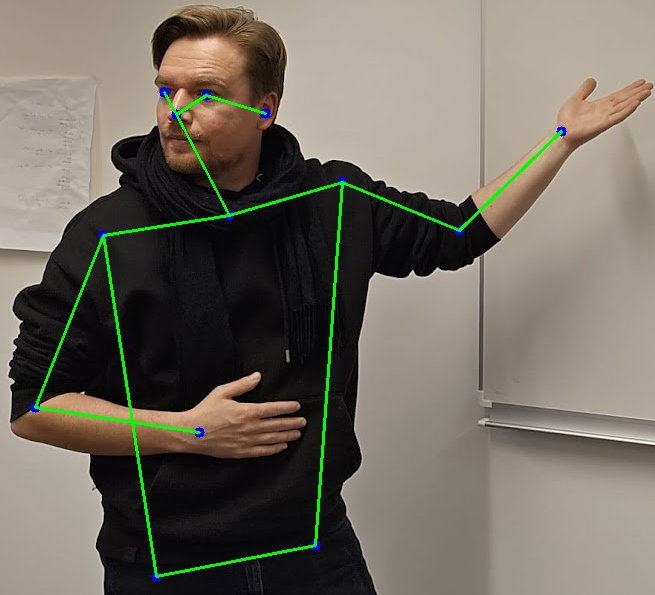}
        \caption{Hand gesture}
        \label{fig:hand_gesture}
    \end{subfigure}\hfill
    \begin{subfigure}[b]{0.35\textwidth}
        \centering
        \includegraphics[width=\textwidth]{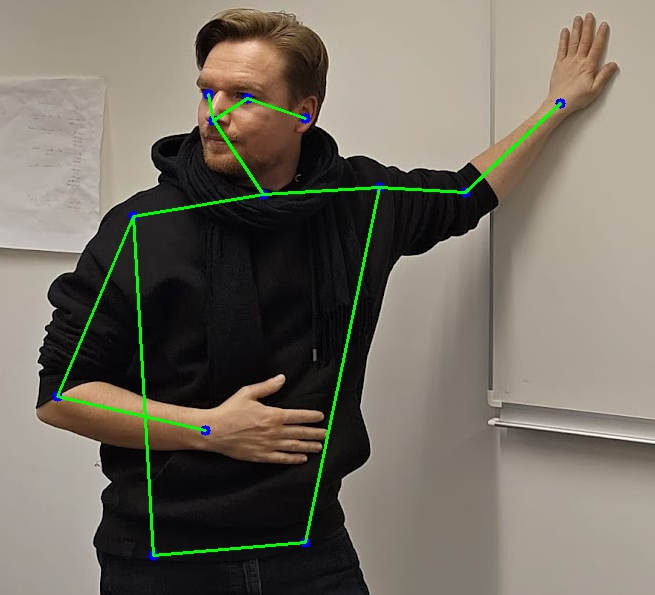}
        \caption{No gesture}
        \label{fig:non_hand_gesture}
    \end{subfigure}
    \caption{Examples of different gesture intensities with similar extracted skeletons. In \ref{fig:hand_gesture}, the person is pointing, representing a specific hand gesture. In \ref{fig:non_hand_gesture}, the person is leaning on the whiteboard, which is not considered gesturing.}
    \label{fig:gestures}
\end{figure}

\section{Dataset and Labeling}

The study utilized the TALIS dataset~\cite{Klieme2019TALISVideostudie}, which includes 135 videos from German schools and represents the German educational system. Each 90-minute video focuses on teaching quadratic equations to 9th and 10th graders. Our study used video data from 46 teachers in the TALIS dataset.

From these videos, three 30-second segments with frontal presentations were extracted, forming the training and validation sets. Only segments without excessive camera movement, where the teacher remained visible, were included. Audio was available for human raters. Segments from 9 teachers were used for validation, and those from 37 teachers for training. To prevent data leakage and memorization, no segments from the same teacher were used in both sets.

For the perceived distance and gesture intensity regressors, 3056 frames were labeled, divided into 2451 for training and 605 for validation. These frames were rated by three trained raters, resulting in an ICC(2,3) of 0.683 for proximity and 0.885 for gesture intensity.

In the context of this study, a 'hand gesture' refers to deliberate movements or signals made by the hand that are intended to communicate specific messages or emotions, as shown in Figure \ref{fig:hand_gesture}. In contrast, a 'non-hand gesture' involves hand movements that are more incidental or casual, not aimed at communication, as illustrated in Figure \ref{fig:non_hand_gesture}. The Figure \ref{fig:gestures} demonstrates the difference in gesture intensity even with similar extracted skeletons.

Perceived distance is evaluated based on the physical space and intervening objects between the teacher and students. Objects like desks placed between the teacher and students can increase the perceived distance, creating a sense of separation. Conversely, the absence of such barriers can lead to a perception of a smaller, more engaging distance.

In a similar approach, 400 30-second video segments (321 for training and 79 for validation) were used for the NVI model. The ICC(2,3) for the NVI score was 0.684, reflecting a substantial level of rater agreement.

\begin{figure}[htbp]
    \centering
    \begin{subfigure}[t]{0.32\textwidth} 
        \centering
        \includegraphics[width=\textwidth]{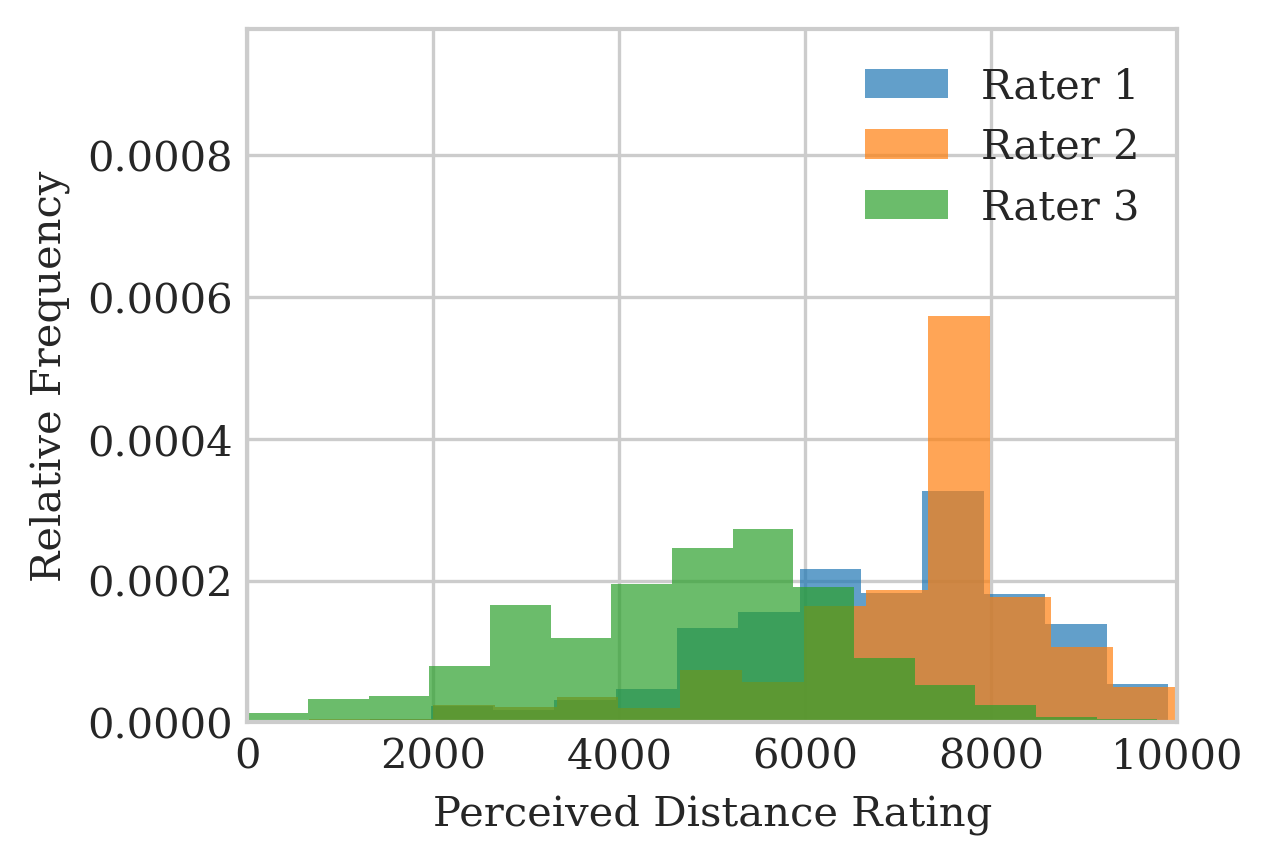}
        \caption{Perceived distance\\rating}
        \label{fig:distance}
    \end{subfigure}
    \hfill 
    \begin{subfigure}[t]{0.32\textwidth} 
        \centering
        \includegraphics[width=\textwidth]{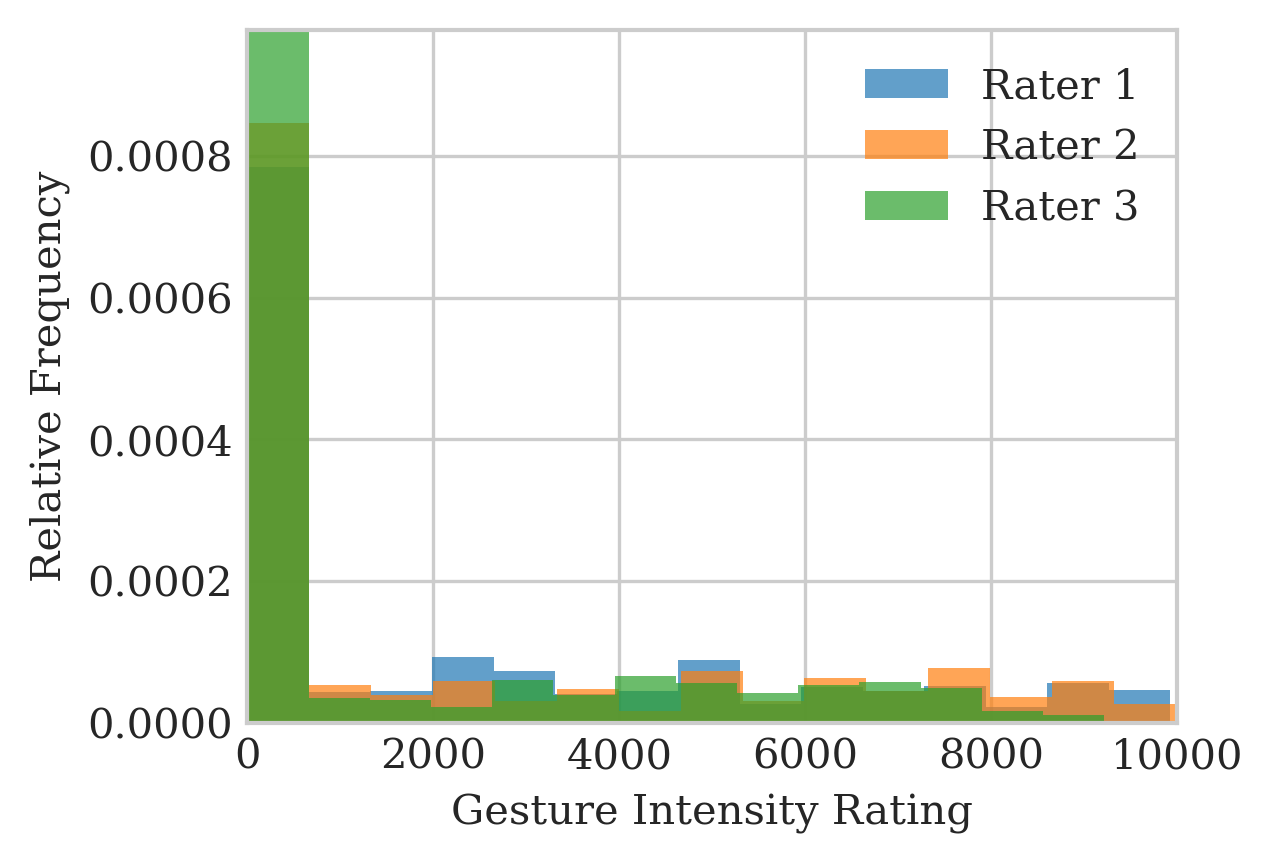}
        \caption{Gesture intensity rating}
        \label{fig:gesture}
    \end{subfigure}
    \hfill 
    \begin{subfigure}[t]{0.32\textwidth} 
        \centering
        \includegraphics[width=\textwidth]{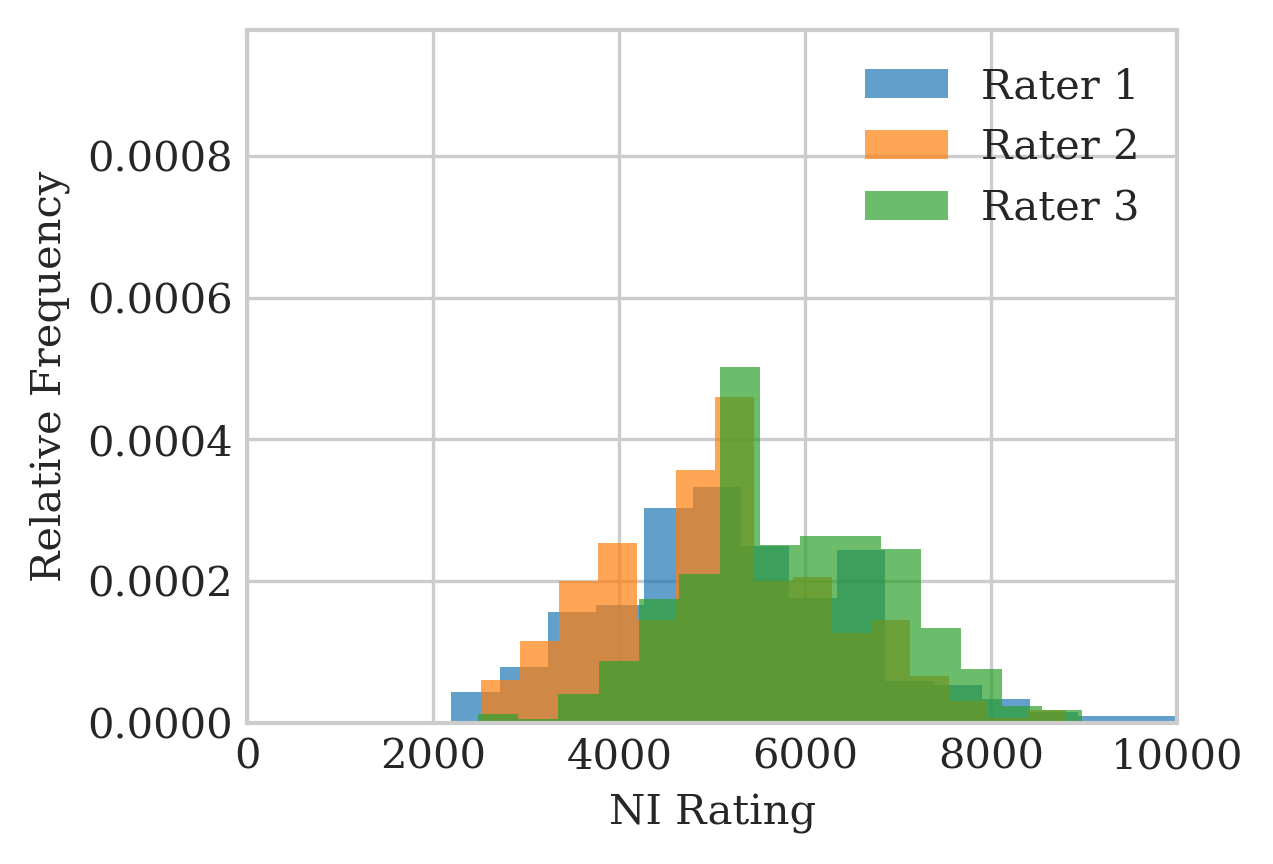}
        \caption{NVI rating}
        \label{fig:immediacy}
    \end{subfigure}
    \caption{Histograms showing the distribution of ratings for perceived distance, gesture intensity, and nonverbal immediacy.}
    \label{fig:histograms}
\end{figure}

Histograms of all datasets are presented in Figure \ref{fig:histograms}.

\section{Methodology}
\label{sec:methodology}

\begin{figure}[htbp]
  \centering
  \includegraphics[width=0.82\linewidth]{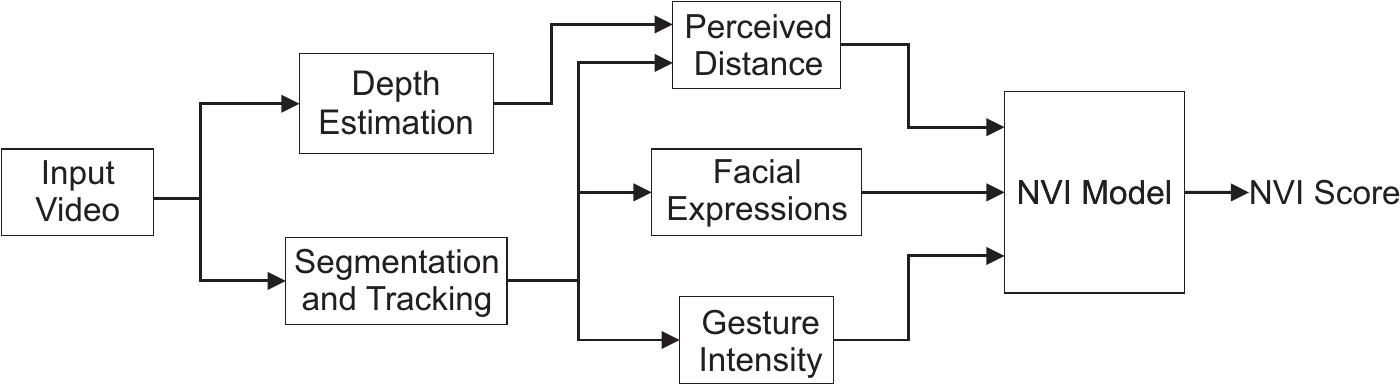}
  \caption{
Pipeline overview: The model takes RGB videos as input, performing tracking, segmentation, and depth estimation. We extract facial expressions and gesture intensity from the teacher's segmentation masks. To estimate perceived distance, we use estimated depth images and segmentation masks from both the teacher and students. These features - facial expressions, gesture intensity, and perceived distance - converge in the NVI model to calculate the NVI score.}
  \label{fig:pipeline}
\end{figure}

In this section, we describe the methods used to measure NVI from teachers in video recordings. NVI estimation is approached as a regression task involving facial expressions, gesture intensity, and perceived distance. We extract facial expressions and gesture intensity from the teacher's segmentation masks. To assess perceived distance, we use estimated depth images and segmentation masks of the teacher and students. These elements are combined, as shown in Figure \ref{fig:pipeline}, to calculate the NI Score from RGB video inputs.

\subsubsection{Tracking and Segmenting}
In each video, the teacher's initial position is manually identified in the first frame. The subsequent tracking of the teacher throughout the video is conducted using Segment and Track Anything~\cite{cheng2023segment}. This involves estimating the semantic segmentation of each frame with the prompt "human"~\cite{kirillov2023segment}. This approach not only segments the teacher but also the students. The segmented data of both teachers and students are later utilized in the gesture intensity and proximity regressors.

\subsubsection{Gesture Intensity Regressor Model}
For the gesture intensity analysis, we developed a specialized model based on a fine-tuned ResNet architecture~\cite{he2016deep}. It extends the standard ResNet18 model, pretrained on ImageNet~\cite{ILSVRC15}, to suit our specific task. The model's input is a masked RGB image of the teacher, obtained from semantic segmentation. It processes this input through ResNet18's layers, followed by additional fully connected layers that adapt the output to our dataset's requirements.

\subsubsection{Perceived Distance Regressor Model}
The perceived distance regressor model mirrors the architecture used in the gesture intensity regressor, based on a fine-tuned ResNet18 architecture~\cite{he2016deep}. The input for this model, consists of a three-channel input: depth image, a mask of the teacher, and a combined mask of all students in a single channel. The depth image was estimated from the RGB frame using~\cite{oquab2023dinov2}. This depth channel provides spatial context, while the masks isolate the subjects of interest, enabling accurate estimation of physical distances in the classroom setting. This format enables the model to recognize spatial relationships between the teacher and students.

\subsubsection{Facial Expression Analysis}
For analyzing facial expressions, we utilized the state-of-the-art HSEmotion library~\cite{savchenko2021facial}, designed for high accuracy and speed in recognizing and quantifying facial expressions. The output from HSEmotion includes confidence scores for eight emotions: anger, contempt, disgust, fear, happiness, neutral, sadness, and surprise, providing a comprehensive emotional profile for each frame. Additionally, in cases where the face is not visible in a frame, we store a distinct output to account for the absence of facial expression data.

\subsubsection{Nonverbal Immediacy Model}
The NVI model in our study, while straightforward in its approach, aligns with methods commonly used by educational psychologists for analyzing NVI~\cite{richmond2003development}. It integrates outputs from gesture intensity and distance regressors, along with facial expression analysis, into a 10-dimensional feature vector. This vector thus combines gesture intensity, distance, and eight emotion confidence ratings.

The integration process weights the feature vector based on the sum of frames with visible faces. The resulting vector feeds into a multilayer perceptron model~\cite{rumelhart1986learning} with three fully connected layers, designed to predict the NVI score. This model reflects the practical approaches in educational psychology, offering a quantifiable measure of various nonverbal aspects.

\section{Experiments}
\label{sec:experiments}

\subsubsection{Regressor Models}
To train the gesture intensity regressor model, we employed the Adam optimizer with a 0.001 learning rate and resized input images to 360x360 pixels. The model was trained with mean squared error (MSE) loss. To ensure the quality of the training subset, we excluded samples with high rater disagreement, defined by a standard deviation of 1600 among ratings. This exclusion reduced the training samples from 2451 to 2121. The median of the three ratings was used as the target for training. For validation, all samples were included, regardless of rater agreement.

Similarly, the perceived distance regressor model was trained using the Adam optimizer with a 0.001 learning rate and 360x360 pixel input images, employing MSE loss. Due to higher rater disagreement, indicated by a standard deviation of 1600, we retained 1604 out of 2451 samples for training. The median of the three ratings was used as the target. For validation, all samples were included for comprehensive model evaluation.

\subsubsection{Nonverbal Immediacy Model}
The NVI model was trained using the Adam optimizer with a learning rate of 0.001 and MSE loss. The model's architecture included three fully connected layers with sizes 300, 100, and 10, respectively. For training, we utilized a majority of the dataset, specifically 316 out of the 321 available samples, selecting based on data quality and consistency. For validation, all samples were included to comprehensively assess the model's performance in various scenarios.

This model being the first, to our knowledge, to analyze NVI using computer vision and given the lack of additional labeled datasets, the model was further validated through an assessment of ecological validity.
An external validation was conducted, to examine the correlation between model-estimated NVI scores and relevant questionnaire data and observer ratings from the TALIS study ~\cite{Klieme2019TALISVideostudie}.

To this end, 220 additional 30-second video clips were extracted from the available video data, applying the same selection criteria as for the original clips. 
Despite efforts to extract the same number of additional clips per video, some teachers had to be excluded, due to the limited availability of sufficient additional video material.

We examined the hypothesized relationships using both the entire dataset (a) and only the additionally identified clips (b). Based on existing NVI-literature~\cite{witt2004meta,liu2021does}, we anticipated positive correlations between the model-estimated NVI scores and student-reported interest in mathematics (H1a + b), cognitive activation (H2a + b), and perceived teacher enthusiasm (H3a + b). Additionally, we hypothesized a positive correlation between NVI scores and teachers’ socio-emotional support behaviors, focusing on the domain of encouragement and warmth (H4a + b), as rated in regular intervals by trained observers during the TALIS-study.
For detailed information regarding the rating process and the scales employed in the TALIS-study see ~\cite{Klieme2019TALISVideostudie}.
To test the hypotheses, a series of Pearson correlations was computed (with false discovery rate adjustments for multiple testing). Questionnaire data and NVI scores were aggregated at the teacher level. For correlations with socio-emotional support ratings, mean aggregates were calculated at the video level.

\begin{table}[h]
\centering
\caption{ICC comparison between human raters and the model on the validation dataset. The ICC values are different from those for the entire dataset as they are estimated solely based on the validation subset.}\label{tab:icc_comparison}
\setlength{\tabcolsep}{6pt}
\begin{tabular}{|l|l|}
\hline
Rater Combination & ICC \\
\hline
Rater0, Rater1, Rater2 & 0.648 \\
Rater0, Rater1, Model & 0.611 \\
Rater0, Model, Rater2 & 0.621 \\
Model, Rater1, Rater2 & 0.613 \\
Rater0, Rater1, Rater2, Model & 0.690 \\
\hline
\end{tabular}
\end{table}

\section{Results}
\label{sec:results}

\subsubsection{Regressor Models}
The gesture intensity regressor model showed a Pearson correlation of 0.84, indicating a high level of predictive accuracy. When used as a binary classifier through thresholding, it achieved an accuracy of 84.9\%. 

The perceived distance regressor model demonstrated a Pearson correlation of 0.55, indicating a moderate positive relationship between predicted values and ground truth. 

\subsubsection{Nonverbal Immediacy Model}
The NVI model exhibited a Pearson correlation of 0.44 with the ground truth based on human raters, indicating a moderate correlation.

As the first to measure NVI computationally, direct comparisons are challenging. To further evaluate the model, we integrated the model-based NVI ratings those of the human raters by calculating median values. This approach allowed for the assessment of bivariate correlations between individual ratings and this new ground truth. Significant and strong positive correlations were found between the median rating and each individual rater, as well as the model: the correlations were 0.74 ($p < .01$) for the first rater, 0.68 ($p < .01$) for the second rater, 0.66 ($p < .01$) for the third rater, and 0.69 ($p < .01$) for the model.

Additionally, as shown in Table \ref{tab:icc_comparison}, replacing one human rater with our model does not significantly drop the ICC, and including the model alongside three human raters slightly improves the ICC. It is important to note that these ICC values differ from those calculated for the entire dataset, as they are estimated solely based on the validation subset.

The results of the correlation analyses in the external validation, considering the full available data set, revealed positive and significant moderate correlations between NVI scores and students' interest in mathematics ($r = .33, p = .03$) (H1a), students' cognitive activation ($r = .32, p = .03$) (H2a), perceived teacher enthusiasm ($r = .31, p = 04$) (H3a), and social-emotional support ($r = .34, p = < .01$) (H4a). 
Using only the additional video data clips, a positive and significant moderate correlation was found between NVI scores and students' interest in mathematics ($r = .45, p < .01$) (H1b) as well as students' cognitive activation ($r = .32, p = .03$) (H2b). 
No significant correlation was observed between NVI scores and perceived teacher enthusiasm ($r = .23, p = .13$) or social-emotional support ($r = .02, p = .86$) using only the additional video data clips.

\section{Discussion and Conclusion}
\label{sec:discussion_and_conclusion}
In this study, we demonstrated the potential of a multimodal computational approach using only RGB cameras to estimate NVI from 30-second video segments in educational settings. We developed a gesture intensity regressor, a perceived distance regressor, and an NVI model integrating these outputs with facial expression analysis. Using our labeled data for training and validation, these models were further evaluated with available data from the TALIS study~\cite{Klieme2019TALISVideostudie} to correlate our model's outputs with real-world educational outcomes.

The gesture intensity regressor model showed strong predictive accuracy, effectively processing gesture intensity from RGB images.
However, limitations in processing the broader context of scenes remain, particularly when faced with complex or ambiguous scenarios.
The perceived distance regressor model exhibited moderate performance, highlighting the complexity of measuring perceived distance compared to geometrical distance.
The NVI model demonstrated a moderate correlation with median human ratings. 
Comparing bivariate correlations while including the models’ scores alongside those of the human raters, demonstrated that the model performs comparable to two of the three observers.
ICC analyses showed, that interrater reliabilities remain stable, when replacing one human rater with the model, while including the model alongside three human raters slightly improves ICC values.
These results indicate, that the model can effectively be leveraged, to support human observers, while fully replacing human raters remains more challenging, due to the complexity and context specificity of nonverbal communication.

The external validation showed significant moderate correlations with students' interest in mathematics, cognitive activation, perceived teacher enthusiasm, and socio-emotional support. These findings indicate that the NVI model effectively captures key aspects of nonverbal communication in educational settings, supporting its potential as a tool for enhancing teacher-student interactions. The discrepancies observed between the outcomes of the two analyses, one utilizing the complete data set and the other employing only the additionally extracted video clips, may be attributed to the reduced quantity of available data in the latter case. The reduced amount of data may limit the representativeness of the analyzed clips for the overall teaching behavior. Moreover, the corresponding clips were not missing at random, with some teachers being excluded completely.

However, we also recognize the challenges encountered. The sensitivity of the data limited the size of our datasets and restricted public availability, making direct comparisons with other studies challenging. Additionally, the variability in video recording conditions across different classrooms, such as camera placement and angles, posed substantial challenges, requiring our models to be adaptable and robust.

Future work will incorporate additional modalities like gaze tracking to enrich understanding of nonverbal communication. While our current study integrated multiple nonverbal modalities, future research will focus on developing new architectures to enhance and support these modalities. Additionally, we will investigate the comprehensive integration of the NVI model and its modalities with scores and relevant questionnaire data and observer ratings from the TALIS study~\cite{Klieme2019TALISVideostudie}. 

In conclusion, our study provides a foundational framework for future research in enhancing teacher-student interactions through a deeper understanding of NVI. It opens promising pathways for educational research and practice, offering a more comprehensive understanding of nonverbal communication between teachers and students.

\section*{Acknowledgments}
We gratefully acknowledge funding by the Deutsche Forschungsgemeinschaft (DFG, German Research Foundation) under Germany’s Excellence Strategy – EXC 2002/1 “Science of Intelligence” – project number 390523135.

%
%
%
%
%


\end{document}